\documentclass[10pt,twocolumn,letterpaper]{article}
\usepackage[accsupp]{axessibility}  

\usepackage{iccv}
\usepackage{times}
\usepackage{epsfig}
\usepackage{graphicx}
\usepackage{amsmath}
\usepackage{amssymb}

\newcommand{\squishlist}{
	\begin{list}{$\bullet$}
		{ \setlength{\itemsep}{0pt}
			\setlength{\parsep}{1pt}
			\setlength{\topsep}{1pt}
			\setlength{\partopsep}{0pt}
			\setlength{\leftmargin}{1.5em}
			\setlength{\labelwidth}{1em}
			\setlength{\labelsep}{0.5em} } }
\newcommand{\squishend}{\end{list} 
}

\usepackage[pagebackref=true,breaklinks=true,letterpaper=true,colorlinks,bookmarks=false]{hyperref}

\iccvfinalcopy 

\def\iccvPaperID{5320} 

\ificcvfinal\fi

\begin{document}

\title{\vspace{-0.2in} Panoramas from Photons}


\author{
  Sacha Jungerman$^\dagger$\\
  {\tt\small sjungerman@wisc.edu}\and
  Atul Ingle$^\S$\\
  {\tt\small ingle2@pdx.edu}\and
  Mohit Gupta$^\dagger$\\
  {\tt\small mohitg@cs.wisc.edu}\and
  {$^\dagger$University of Wisconsin-Madison}\quad
   {$^\S$Portland State University}}


\maketitle

\renewcommand*{\thefootnote}{$\ddagger$}
\setcounter{footnote}{1}
\footnotetext{This research was supported in part by an NSF CAREER award 1943149, NSF award CNS-2107060 and NSF ECCS-2138471.}

\renewcommand*{\thefootnote}{\arabic{footnote}}
\setcounter{footnote}{0}

\ificcvfinal\thispagestyle{empty}\fi

\begin{abstract}
\vspace{-0.1in}
Scene reconstruction in the presence of high-speed motion and low illumination is important in many applications such as augmented and virtual reality, drone navigation, and autonomous robotics. Traditional motion estimation techniques fail in such conditions, suffering from too much blur in the presence of high-speed motion and strong noise in low-light conditions. Single-photon cameras have recently emerged as a promising technology capable of capturing hundreds of thousands of photon frames per second thanks to their high speed and extreme sensitivity. Unfortunately, traditional computer vision techniques are not well suited for dealing with the binary-valued photon data captured by these cameras because these are corrupted by extreme Poisson noise. Here we present a method capable of estimating extreme scene motion under challenging conditions, such as low light or high dynamic range, from a sequence of high-speed image frames such as those captured by a single-photon camera. Our method relies on iteratively improving a motion estimate by grouping and aggregating frames after-the-fact, in a stratified manner. We demonstrate the creation of high-quality panoramas under fast motion and extremely low light, and super-resolution results using a custom single-photon camera prototype. For code and supplemental material see our \href{https://wisionlab.com/project/panoramas-from-photons/}{project webpage}.
\vspace{-0.1in}
\end{abstract}

\vspace{-0.05in}
\section{Introduction}\label{sec:intro}
\vspace{-0.05in}

\begin{figure*}[!t]
    \centering \includegraphics[width=0.95\textwidth]{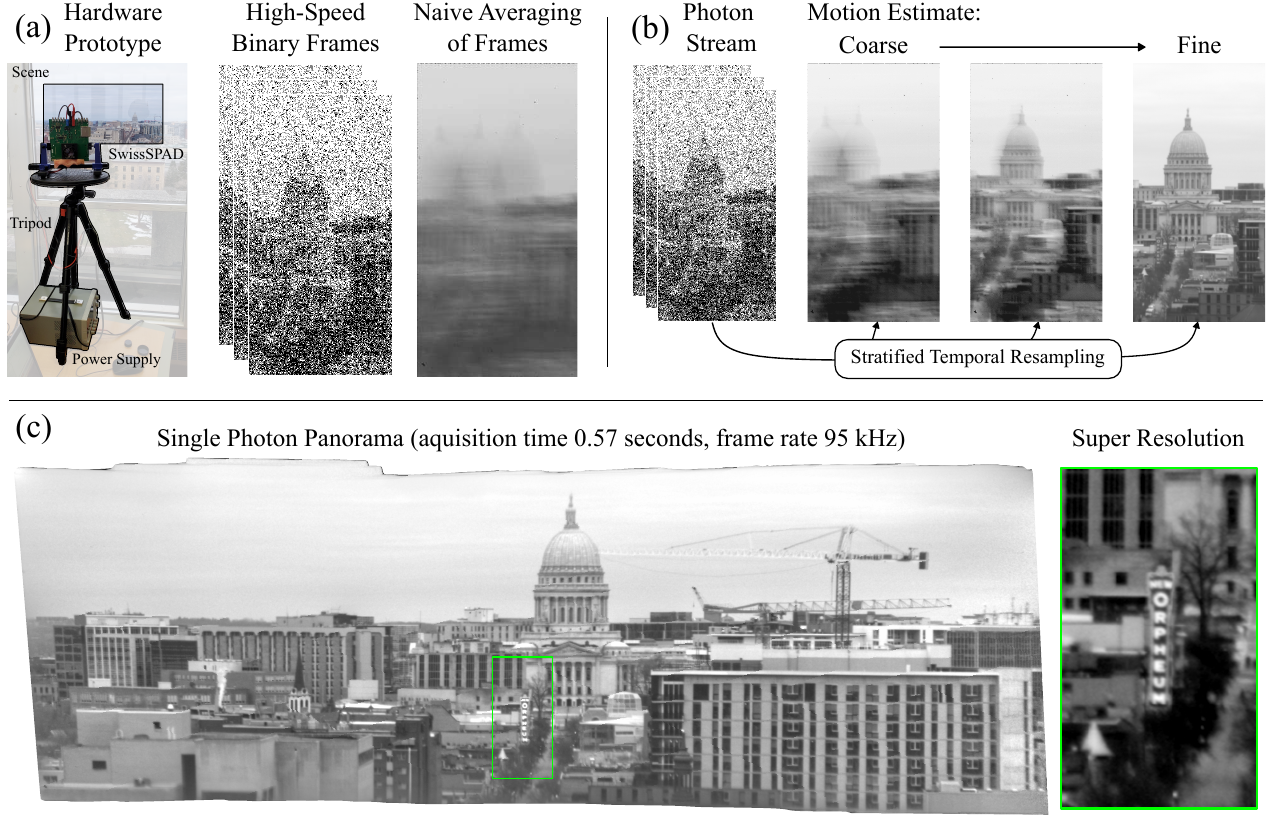}
    \caption{\textbf{Single photon panoramas:}
    (a) Our single-photon camera prototype can capture binary frames at $\sim 100,000$ frames per second. 
  Conventional processing techniques that average the raw image frames struggle due to extreme motion blur. (b) Our proposed method recovers a high-quality scene reconstruction by iteratively refining a motion estimate and re-aggregating the raw photon data. (c) An example high-speed panorama reconstructed from single-photon frames in merely half a second total capture time. 
  \label{fig:teaser}}
  \vspace{-0.1in}
\end{figure*}

Accurate recovery of motion from a sequence of images is one of the most fundamental tasks in computer vision, with numerous applications in robotics, augmented reality, user interfaces, and autonomous navigation. When successfully estimated, motion information can be used to locate and track the camera or different objects in the scene~\cite{comaniciu_kernel-based_2003}, perform motion-aware video compression~\cite{ISO14496} or stabilization~\cite{lee_video_2009}, relate multiple sensors, merge information across different viewpoints, and even reconstruct city-scale 3D models using only images from the web~\cite{agarwal_building_2009,schonberger_structure--motion_2016,martin-brualla_nerf_2021}.

Image sequences can be used to estimate different kinds of motion ranging in complexity and degrees of freedom, from global motion models, such as simple translations, projective warps, or 3D (6-DoF) camera pose, to non-rigid, local motion models such as optical flow. However, regardless of the motion model, traditional methods cannot recover motion that is simply too fast for the camera to capture. This is especially challenging when capturing scenes in low-light conditions---the camera will compensate by increasing the exposure, thereby introducing motion blur, as seen in Fig.~\ref{fig:teaser}(a), or increasing the gain (ISO), thereby introducing noise \cite{hasinoff_noise-optimal_2010}.
Fundamentally, the image degradation associated with faster motion or a darker scene causes traditional motion estimation methods to fail.


One way to handle fast motion is by using specialized high-speed cameras. However, such cameras are not only bulky and costly but also suffer from extremely low signal-to-noise ratio due to both low signal values and high readout noise, at least an order of magnitude higher than conventional CMOS cameras\footnote{For example, the \href{https://www.phantomhighspeed.com/products/cameras/ultrahigh4mpx/v2640}{Phantom v2640 has read noise up to $58e^-$}.}. This requires the scenes to be well illuminated, often in a controlled setting, further limiting their scope and widespread adoption. 


Fortunately, there is an emerging class of sensors called single-photon cameras, which are capable of high-speed imaging in low-light conditions. Single-photon cameras based on single-photon avalanche diode (SPAD) technology \cite{bruschini2019single} provide extreme sensitivity, are cheap to manufacture, and are increasingly becoming commonplace, recently getting deployed in consumer devices such as iPhones. The key benefit of SPADs is that they do not suffer from read-noise, enabling captures at hundreds of thousands of frames per second even in extremely low flux, while being limited only by the fundamental photon noise. 

Although single-photon cameras can capture scene information at unprecedented sensitivity and speed, each individual captured frame is binary valued: a pixel is ``on" if at least one photon is detected during the exposure time and ``off" otherwise. This binary imaging model presents unique challenges. Traditional image registration techniques rely on feature-based matching, or direct optimization using differences between pixel intensities, both of which rely on image gradients to converge to a solution. Individual binary images suffer from severe noise and quantization (only having 1-bit worth of information per pixel), and are inherently non-differentiable, making it challenging, if not impossible, to apply conventional image registration and motion estimation techniques directly on binary frames. Aggregating sequences of binary frames over time increases signal (Fig.~\ref{fig:teaser}(a)) but comes at the cost of potentially severe motion blur, creating a fundamental noise-vs-blur tradeoff. 

We present a technique capable of estimating rapid motion from a sequence of high-speed binary frames captured using a single-photon camera. Our key insight is that these binary frames can be aggregated in post-processing in a \emph{motion-aware manner} so that more signal and bit-depth are collected, while simultaneously minimizing motion blur. As seen in Fig.~\ref{fig:teaser}(b), our method iteratively improves the initial motion estimate, ultimately enabling scene reconstruction under rapid motion and low light and conditions. 

\noindent {\bf Scope and Capabilities:} We demonstrate the recovery of global projective motion (homography), enabling the capture of high-speed panoramas with super-resolution and high dynamic range capabilities. As shown in Fig.~\ref{fig:teaser}(c), our algorithm can reconstruct a high-quality panorama, captured in less than a second over a wide field-of-view, while simultaneously super-resolving details such as text from a long distance ($\sim 1300$ m). The ideas presented in this paper could also be used to enhance recent one-shot local motion compensation work~\cite{ma_quanta_2020,seets_motion_2020,iwabuchi_image_2021} as they are complementary.



\noindent {\bf Limitations:} Although single-photon camera technology is rapidly evolving, today's SPAD arrays suffer from limitations such as low fill-factors, low spatial resolution, and lack of high-quality color filters. This limits the visual quality of the experimental results shown here. Fortunately, given the trend towards higher resolution SPAD arrays \cite{morimoto2020megapixel} and the increasing commercial availability of this technology \cite{morimoto20213}, these are not fundamental limitations.





\section{Related Work}\label{sec:related-works}
\vspace{-0.05in}
\noindent \textbf{Image Stitching:} Merging multiple images together to create a large cohesive image, referred to as a panorama or mosaic, is a classical problem in computer vision. It consists of two main steps, namely image registration, and merging. To register images, most approaches either rely on computing image features, such as SIFT features~\cite{lowe_distinctive_2004}, or direct optimization of the warps, such as the Lucas-Kanade algorithm~\cite{lucas_iterative_1981} or variants thereof~\cite{baker_lucas-kanade_2004-1}. Once features are extracted, it is possible to match them between images to compute the warps that relate one image to another~\cite{brown_automatic_2007}. More recent techniques use learning-based methods to extract features~\cite{liu_image_2018} allowing them to match cross-domain images such as satellite and map images~\cite{zhao_deep_2021}. Unfortunately, the presence of extreme Poisson noise causes traditional stitching approaches to fail for high-speed binary frames. 

\smallskip
\noindent \textbf{Structure from Motion (SfM):} SfM techniques estimate both the 3D geometry of the environment and the location of the camera simultaneously. Most SfM pipelines (e.g., COLMAP~\cite{schonberger_structure--motion_2016}) use feature-based approaches to match frames. Others have extended the optimization-based approaches of Lucas-Kanade to 3D pose estimation~\cite{baker_lucas-kanade_2004,engel_direct_2016}. While these methods can be robust to some types of noise, such as Gaussian, salt and pepper, and speckle~\cite{routray_analysis_2017}, they all rely on computing image gradient either as part of the optimization process or as part of the feature extraction process and thus are not well suited for noisy and quantized high-speed images such as binary images. 

\smallskip
\noindent \textbf{Burst Photography and Denoising:} Techniques that remove image noise can be used as a pre-processing step to aid the feature matching or direct optimization process. For instance, blind denoisers~\cite{dabov_image_2007} or state-of-the-art deep video denoising networks~\cite{tassano_fastdvdnet_2020} could be used. The burst processing of binary frames is also possible~\cite{ma_quanta_2020} but relies on motion compensation to denoise images which can be expensive as it requires computing optical flow on each frame. For many scenes, computing optical flow is unnecessary as motion might be primarily dominated by ego-motion. We propose an iterative approach that refines a global motion estimate and enables high-quality scene recovery while being computationally less burdensome than optical flow approaches that use patch-based processing.

\smallskip
\noindent \textbf{Event-based Processing:} Event cameras (dynamic vision sensors) superficially resemble high-speed binary frames: they produce high frequency, low-bit depth observation of a scene and have been used for fast tracking~\cite{kim_simultaneous_2014} and odometry~\cite{hidalgo-carrio_event-aided_2022}. However, event cameras suffer from high sensor noise and event clutter caused by camera motion~\cite{gallego_event-based_2022} leading many works to use a fusion approach, combining events with conventional camera frames. Our method focuses on intensity frames, whether captured from a single photon camera or other sensing modality.

\section{Image Formation via Virtual Exposures}\label{sec:keyidea}

\begin{figure}[!t]
    \centering \includegraphics[width=1.0\columnwidth]{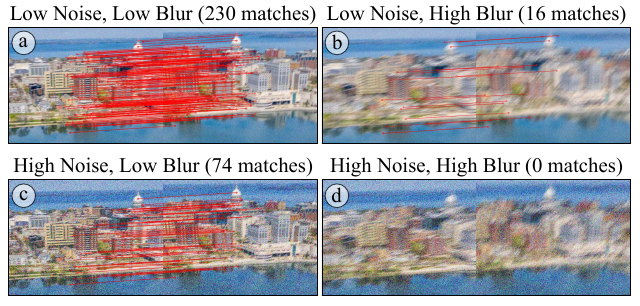}
    \caption{\textbf{Registration Accuracy vs. Noise and Blur:} Conventional feature matching techniques, such as exhaustively matched SIFT features shown here, work well with low noise, and low motion blur frames. The number of successful feature matches drops off at higher noise and blur levels.
    \label{fig:feature-failure}}
    \vspace{-0.1in}
\end{figure}

Consider a series of images captured by a camera as the scene and/or camera undergoes motion. Suppose our goal is to register a pair of consecutive images, under a given motion model (local or global). In ideal imaging conditions (sufficient light, relatively small motion), conventional motion estimation and registration techniques perform robustly. This is demonstrated in Fig.~\ref{fig:feature-failure} (a), where a SIFT-based feature matching technique is able to find reliable matches across images. However, in settings involving low-light and rapid motion, the number of successful feature matches drop, resulting in erroneous motion estimation. This is due to the fundamental noise-vs-blur trade-off---the captured images either have strong noise or large motion blur, depending on the exposure length used, both of which prevent reliable feature detection and image registration (Fig.~\ref{fig:feature-failure} (b-d)). More generally, this trade-off limits the performance of several computer vision and imaging techniques that require motion estimation across a sequence of images~\cite{hasinoff_burst_2016,liu_mba-vo_2021}. Is there a way to overcome this trade-off? \smallskip



\noindent \textbf{Mitigating Blur-Noise Trade-off via Virtual Exposures:} Conventional cameras integrate the scene's radiance during an exposure, and produce a single image. Under the challenging conditions described above, this image may be too noisy or blurred, depending on the exposure duration. Suppose instead, we were able to record the arrival time of each photon during an exposure, creating a \emph{3D photon cube} ($x$ and $y$ spatial dimensions, and an extra photon arrival time dimension)~\cite{fossum2005sub,fossum2011quanta}. This information is richer than what a conventional camera image can afford us, but what can we do with it? While we can reconstruct a conventional camera image by simply summing over the time slices of the photon cube, we can combine this photon data in multiple ways post hoc. We could apply arbitrary transformations to each time slice before collapsing it into one or more final images. We refer to this idea of aggregating photon information after-the-fact as \textit{virtual exposures}. In contrast, once a conventional image has been captured, undoing the effect of motion artifacts is severely ill-posed. \smallskip


\begin{figure*}[!t]
    \centering \includegraphics[width=1.0\textwidth]{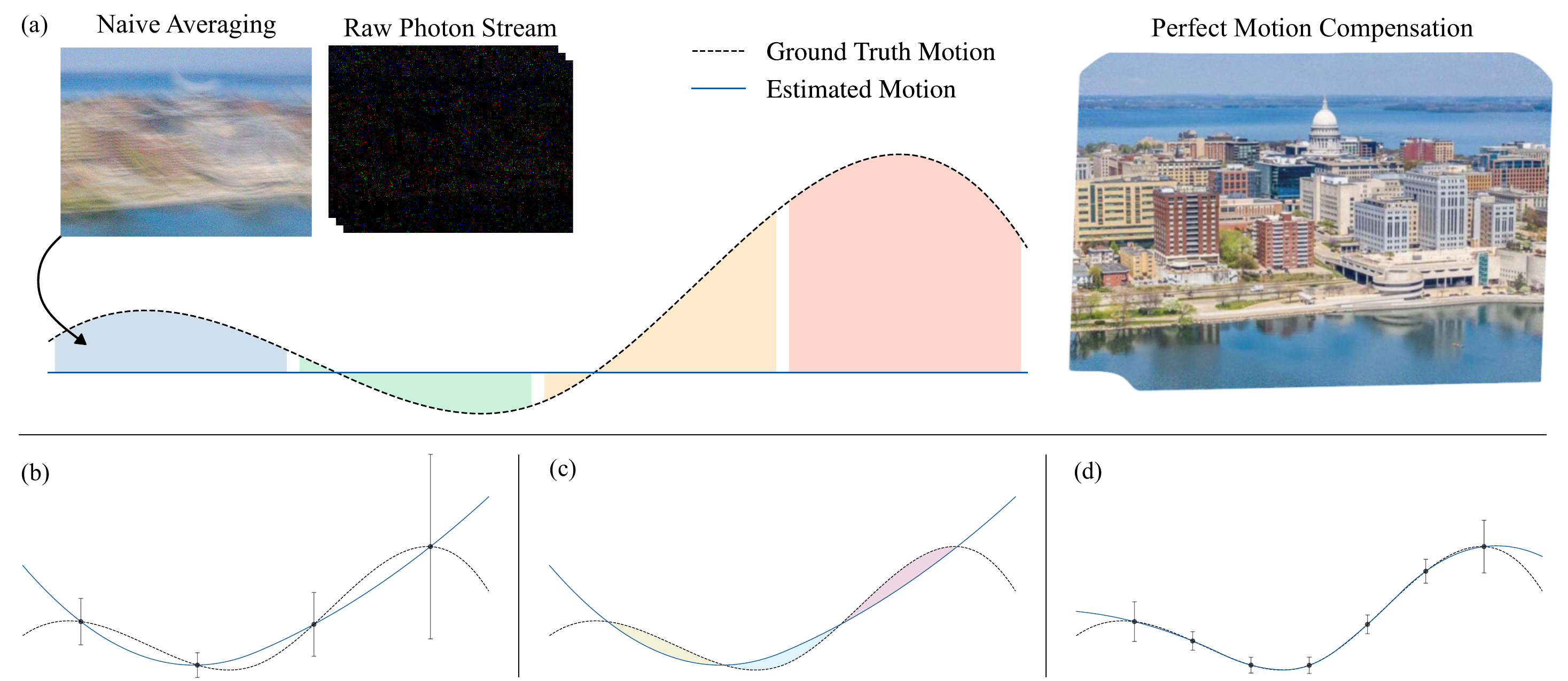}
    \caption{\textbf{Motion Estimation using Stratified Temporal Re-Sampling:} Perfect scene reconstruction can be achieved if the scene motion is precisely known. 
(a) We cannot recover a motion estimate from the raw photon data as it is binary-valued, extremely noisy, and non-differentiable.
An initial motion estimate (blue line) is obtained using locally averaged groups of frames (shaded regions).
(b) This blur causes the registration algorithm to produce noisy motion estimates (black error bars) from which we can update our estimated motion trajectory.
(c) With this new trajectory, the apparent motion is smaller (shaded area), leading to higher-quality virtual exposures. We can also sample new virtual exposures as needed, here we show new frames centered around the midpoints of previous frames. 
(d) These lead to improved motion estimates.
The estimated motion trajectory converges to the true motion over several iterations.
\label{fig:iterated-estimation}} 
\vspace{-0.1in}
\end{figure*}


\noindent \textbf{Stratified Temporal Re-Sampling:} Our key insight, enabled by the concept of virtual exposures, is that we can compensate for motion at the level of individual photon arrivals to create high-fidelity aggregate frames, which in turn can be used to further refine the motion estimates. Virtual exposures are created by re-sampling the photon-cube post-capture, allowing arbitrary, fluid, and even overlapping exposures, enabling us to resolve higher speed motion.


An abstract example of this concept is illustrated in Fig.~\ref{fig:iterated-estimation}. We start with an initial set of virtual exposures, which are simply aggregate frames with no motion compensation akin to a sequence of short exposures from a conventional camera. From these, we estimate a coarse motion trajectory using an off-the-shelf motion model. Although conventional techniques will output potentially erroneous estimates, we propose an iterative approach, where these motion estimates are used to spatiotemporally warp the underlying photon data and re-combine it into less blurry images. This is repeated to create additional virtual exposures until convergence, resulting in improved motion estimates. \smallskip


\noindent {\bf How to Capture Photon Cubes?} Thus far, we assumed having access to a continuous-time stream of photons that contains precise timing and location information. In practice, we approximate this photon stream with a camera capable of high-speed temporal sampling. While several high-speed sensing technologies exist today, we focus on single-photon avalanche diode (SPAD) sensors. In the following, we describe their unique image formation model that enables high-speed photon-level sensing, which can emulate virtual exposures whose signal-to-noise ratio (SNR) is limited only by the fundamental limits of photon noise. \smallskip



\noindent \textbf{SPAD Image Formation Model:} For a static scene with a radiant flux (photons/second) of $\phi$, during an exposure time $\tau$, the probability of observing $k$ incident photons on a SPAD camera pixel follows a Poisson distribution:
\vspace{-0.05in}
\begin{equation}
  P(k) = \frac{(\phi \tau)^k e^{-\phi \tau}}{k !}.
\end{equation}
\vspace{-0.05in}
After each photon detection, the SPAD pixel enters a dead time during which the pixel circuitry resets. During this dead time, the SPAD cannot detect additional photons. The SPAD pixel output during this exposure $\tau$ is binary-valued and follows a Bernoulli distribution given by\footnote{Source of noise such as dark counts and non-ideal quantum efficiency can be absorbed into the value of $\phi$.}:

\vspace{-0.175in}

\begin{equation}
  P(k=0) = e^{-\phi \tau}, \qquad P(k=1) = 1-e^{-\phi \tau}. \label{eq:binary-imgs}
\end{equation}

\smallskip
\noindent \textbf{Emulating Virtual Exposures:} Given $n$ binary observations $B_i$ of a scene, we can capture a virtual exposure using the following maximum likelihood estimator~\cite{yang2011bits}:
\begin{equation}
  \widehat{\phi} = - \frac{1}{\tau} \ln \left(1- \frac{1}{n} \sum_{i=1}^n B_i \right). \label{eq:flux-mle}
\end{equation}
Different virtual exposures can be emulated by varying the starting index $i$ and the number $n$ of binary frames. The granularity and flexibility of these virtual exposures is limited only by the frame rate of the SPAD array, which reaches up to $\sim 100k fps$, enabling robust motion estimation at extremely fine time scales. Furthermore, SPAD arrays have negligible read noise and quantization noise, leading to significantly higher SNR as compared to conventional images captured over the same exposures.



\begin{figure*}[!t]
    \centering \includegraphics[width=1.0\textwidth]{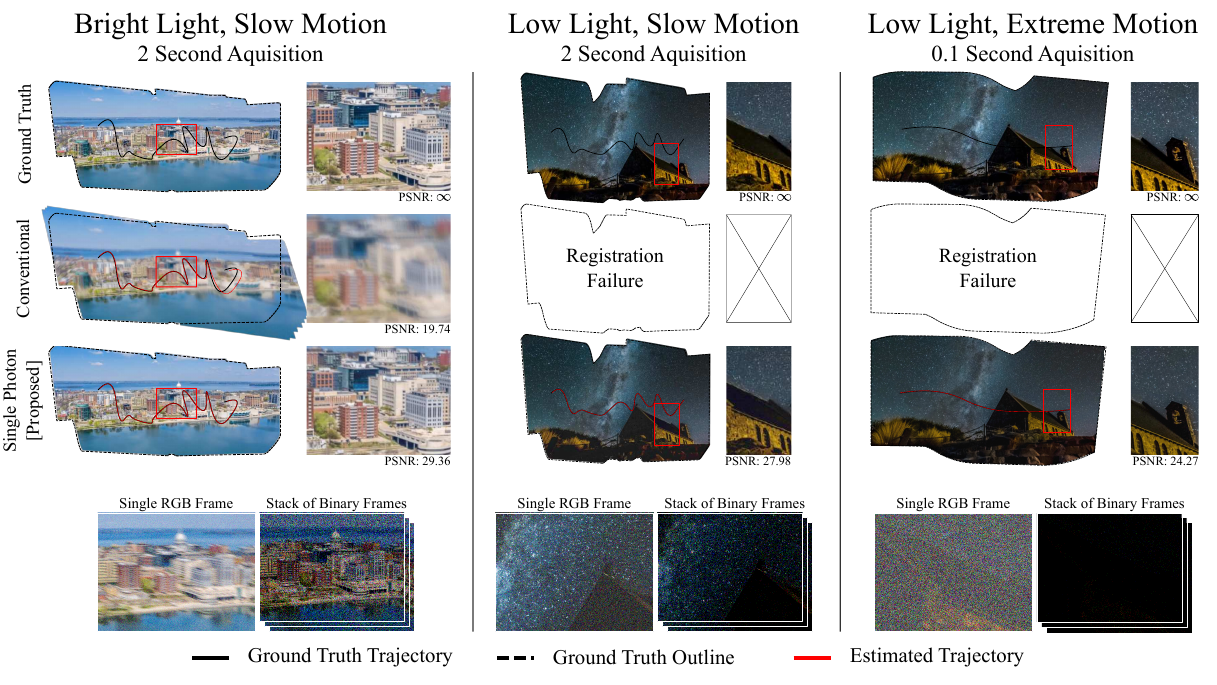}
    \caption{\textbf{Simulated Panoramas:} (left) We show a ground truth panorama, and a zoomed-in section of it, as well as one created with realistic blurry RGB frames and another created under the same conditions using binary frames.
Reliable reconstruction using traditional RGB frames is impossible due to large motion blur, however, we can compensate for fine-grain motion using binary frames resulting in perfect reconstruction. (middle) The baseline fails in low light, this scene is $6 \times$ darker than the left one. 
(right) Our method works even with $120 \times$ less light. The baseline reconstruction fails as each conventional image is dominated by read noise. \label{fig:pano-comparison}}
\vspace{-0.1in}
\end{figure*}

\section{Stratified Estimation of High-Speed Motion}\label{sec:methods}

While the ideas presented in Section~\ref{sec:keyidea} are applicable to a wide range of motion models, we focus on image homographies, a global motion model. We propose a modular technique for homography estimation from photon cube data, which is capable of localizing high-speed motion even in ultra-low light settings. As an example application, we demonstrate panoramic reconstruction from photon cubes by using the estimated homographies to warp binary frames onto a common reference frame. 
Given a temporal sequence of $n$ binary frames $\{B_i\}_{i=1}^n$, we compute and iteratively refine image homographies and the resulting reconstruction through the following steps:



\squishlist
    \item \textbf{Re-sample:} Sample binary frames across the photon cube which will be merged together. 
    \item \textbf{Merge:} Merge the sampled frames using the current per-frame homography estimate.
    \item \textbf{Locate:} Apply an off-the-shelf motion estimation algorithm to the merged frames.
    \item \textbf{Interpolate:} Interpolate the estimated homographies to the granularity of individual binary frames.  
\squishend

With successive iterations of the above method, the homography estimates are refined. Once convergence is reached, the per-frame estimated warps are used to assemble the final panorama.  


\begin{figure*}[!t]
    \centering \includegraphics[width=0.95\textwidth]{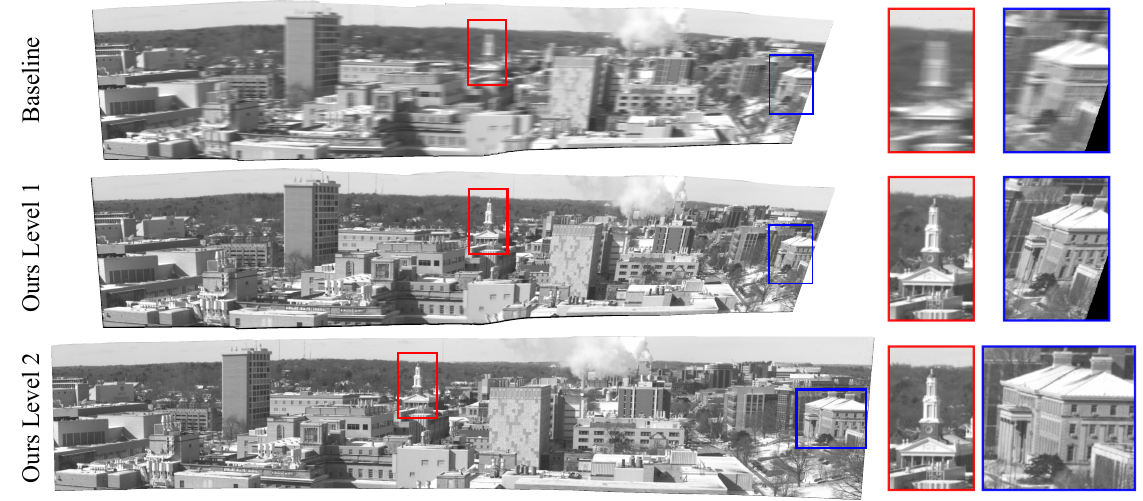}
    \caption{\textbf{Multi-level refinement of panorama:} We show a panorama created by naively averaging adjacent frames, in groups of $1000$ (baseline), and two iterations of our method also using a group size of $m=1000$ (ours, 1 and 2). 
  Blurrier regions such as the tower (red inset) become sharper and the building (blue inset) is reconstructed without distortion after only two iterations. \label{fig:mulilvl}}
  \vspace{-0.15in}
\end{figure*}

\begin{figure}[!t]
    \centering \includegraphics[width=0.95\columnwidth]{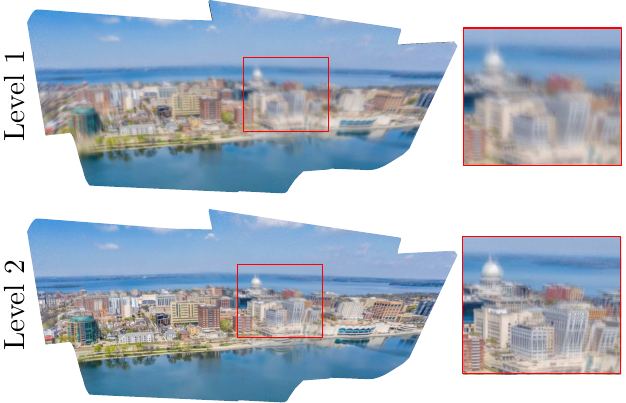}
    \caption{\footnotesize \textbf{Constant number of groups without resampling:} Without adding new virtual exposures at each level, the motion trajectory estimates are unreliable, resulting in blur and ghosting artifacts. \label{fig:constantgroup}}
    \vspace{-0.15in}
\end{figure}

\smallskip
\noindent \textbf{Re-sample:} The entire sequence of binary frames is re-sampled and grouped into subsets that are later aligned and merged. We use midpoint sampling as the grouping strategy. Given a group size of $m$, during the first iteration, we split the $n$ binary frames into $\lfloor n/m \rfloor$ non-overlapping groups. A single frame within each group is chosen to be the reference frame whose warp is later estimated in the ``Locate'' step. Initially, we choose the center frame of each group to be the reference frame. In subsequent iterations, the binary frame sequence is \emph{re-sampled} to create new groups consisting of $m$ frames that are chosen such that they are centered between the previous iteration's groups. This introduces overlapping groups and ensures a progressively denser sampling of the motion trajectory. Fig.~\ref{fig:constantgroup} illustrates what happens if we omit this crucial step---regions where the motion trajectory is more complex exhibit blur and ghosting artifacts.  

This stratified re-sampling approach is crucial for dealing with the motion blur and noise tradeoff. 
The number of frames ($m$) per group plays an important role in dealing with this tradeoff: a larger value of $m$ helps counteract Poisson noise but also causes motion blur.
In practice, if SPAD binary frames are available at $\sim 100$ kHz, setting $m\approx$250-750 achieves high-quality results across different motion speeds and light levels, with higher values better suited for extremely low light, and lower values for fast motion. See supplementary material for details on the asymptotic behavior of this grouping policy, and the impact of the choice of the reference frame for each group.




\smallskip
\noindent \textbf{Merge:} The frames within each group are warped and merged. The warp operation is applied locally within each group. Applying these warps locally with respect to the group's center frame (instead of a global reference frame) is critical; it ensures that the frames within each group only need to be warped by small amounts.
The warped frames are then merged using Eq.~(\ref{eq:flux-mle}) and tone-mapped to sRGB.

\smallskip
\noindent \textbf{Locate:} The pairwise warps between merged frames are estimated using an off-the-shelf method. Any drift introduced in this step is corrected during subsequent iterations.

\smallskip
\noindent \textbf{Interpolate:} The ``Locate'' step estimates homographies across groups of \emph{merged} binary frames. In this step, we interpolate these estimated homography matrices across time to get the fine-scale warps later used to warp \emph{individual binary frames}. A natural way to interpolate homographies is using a geodesic interpolation
~\cite{fragneto_uncalibrated_2012}. In practice, an extended Lucas-Kanade formulation~\cite{baker_lucas-kanade_2004-1} is more robust since it avoids computing matrix inverses and is numerically more stable. We perform cubic interpolation on the eight free parameters, $p_i$, of the $3\times 3$ homography matrix: 
\begin{equation}
    H = \begin{bmatrix}
        1+p_1 & p_3 & p_5\\
        p_2 & 1+p_4 & p_6\\
        p_7 & p_8 & 1
    \end{bmatrix}.
\end{equation}

The resulting interpolated homographies are able to resolve extremely high-speed motion at the granularity of individual binary frames ($\sim 100$ kHz), thus significantly mitigating the noise-blur tradeoff. \smallskip

\begin{figure*}[!ht]
    \centering \includegraphics[width=1.0\textwidth]{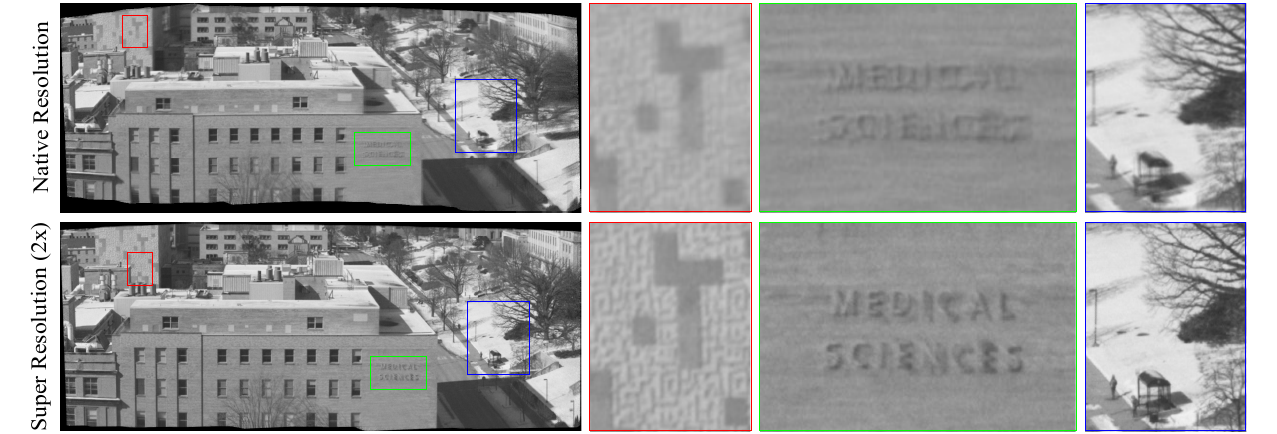}
    \caption{\textbf{Super-resolution on experimental data:} By interpolating homographies over additional virtual exposures, our method can super-resolve the sensor's native resolution ($254 \times 496$) by $2\times$. Details such as text on the building (red, green insets) and finer structures such as tree branches (blue inset) are super-resolved.
    \label{fig:sr}}
    \vspace{-0.1in}
\end{figure*}

\noindent \textbf{Computational Considerations:} Instead of the proposed re-sampling method, one could create virtual exposures centered around each time instance, in a sliding window manner, and register those. This would not only be computationally expensive, as one would need to construct (and extract features from) hundreds of thousands of aggregate frames, but would also produce blurry results as the registration process will be sensitive to the blur introduced in each aggregate frame. The iterative nature of the proposed method allows for progressively better localization as at each iteration, the merged frames are progressively less blurry. This is shown in Fig.~\ref{fig:iterated-estimation}(d) as the error bars on existing points get smaller in the second iteration. Thus, the naive sliding-window approach would also need to be iterated upon, yielding a complexity of $\mathcal{O}(n\cdot m)$, for processing $n$ frames over windows of size $m$ (assuming the number of iterations stays constant). In contrast, our iterative method has an asymptotic runtime of $\mathcal{O}(n/m)$, providing significant speedup over the sliding-window approach.



\section{Experiments and Results}\label{sec:results}



\subsection{Setup: Simulations and Hardware}

We demonstrate our technique in simulation and through real-world experiments using a SPAD hardware prototype. 

\smallskip
\noindent \textbf{Simulation Details:} We simulate a SPAD array capturing a panoramic scene by starting with high-resolution panoramic images downloaded from the internet. We create camera trajectories across the scene such that the SPAD's field of view (FOV) sees only a small portion of the panorama at a time. At each time instant of the trajectory, we simulate a binary frame from the FOV of the ground truth image by first undoing the sRGB tone mapping to obtain linear intensity estimates, and then applying Eq.~(\ref{eq:binary-imgs}) to simulate the binary photon stream. RGB images are simulated by averaging the ground truth linear intensities over a certain exposure and adding Gaussian noise~\cite{hasinoff_noise-optimal_2010}. \smallskip


\noindent \textbf{Hardware Prototype:} For real experiments, we use the SwissSPAD~\cite{ulku_512_2019} to capture binary frames (Fig.~\ref{fig:teaser}(a)). The sensor has a usable resolution of $254 \times 496$ pixels. It does not have micro-lenses, or a color filter array, and the fill factor is $10.5\%$ with $16.8\mu m$ pixel pitch. Despite these limitations, it is capable of capturing binary frames at 100 kHz. 

\noindent \textbf{Implementation Details:} We use OpenCV's registration algorithm based on SIFT and RANSAC homography fitting to match virtual exposures. Our implementation takes roughly ten minutes per iteration to process $100k$ frames. While factors such as resolution and window size ($m$) will affect runtime, our implementation is throttled by the underlying registration algorithm which recomputes features at every level. Further optimizations and feature caching would greatly improve runtime. 


\begin{figure*}[!t]
    \centering \includegraphics[width=1.0\textwidth]{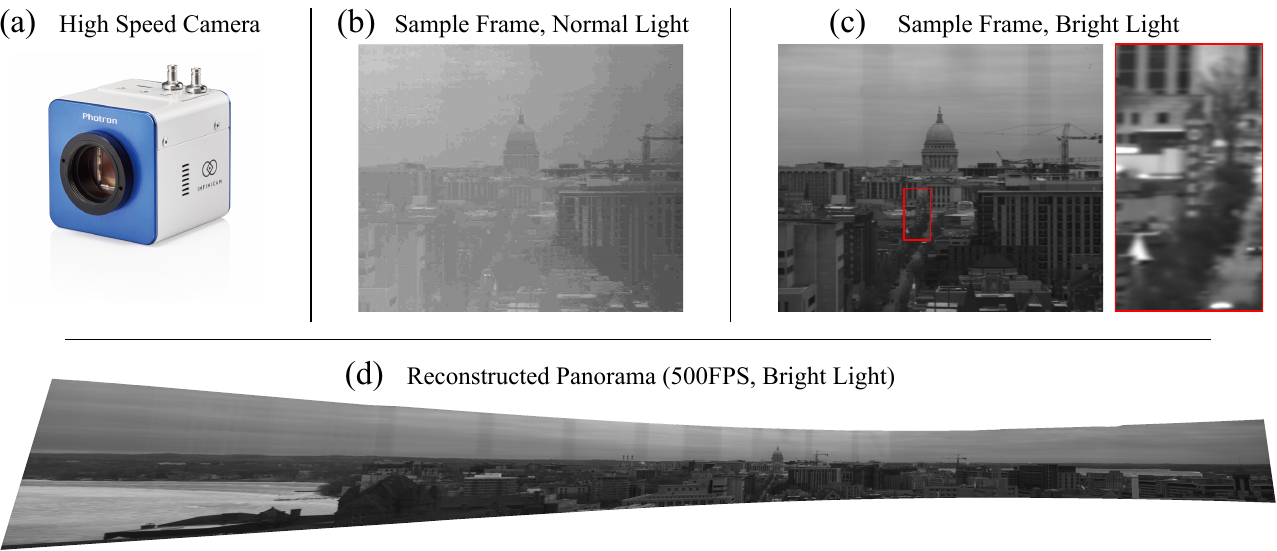}
    \caption{\textbf{Application to conventional high-speed cameras:} (a) We use Photron's Infinicam as our high-speed camera mounted with similar optics as our SPAD camera prototype. (b) Individual image frames from the commercial high-speed camera are extremely noisy and show compression artifacts even when capturing frames at $\sim 2\times$ slower motion and running at $500$fps.  Panorama reconstruction using such frames fails due to a lack of reliable feature matches across frames. 
    (c) Using a lens with a larger aperture ($f/2.8$ instead of $f/5.6$) the frame quality dramatically improves. However, despite the Infinicam's high resolution of $1246 \times 1024$ and slower motion, the text on the sign is still blurry compared to the SPAD camera result in Fig.~\ref{fig:teaser}(c). (d) In this higher flux regime, we can reconstruct a panorama (the vertical band artifacts are reflections from the window), albeit a blurry one as each individual frame is blurry. This demonstrates that while our method works with high-speed cameras, it is ideally suited for single-photon imaging. 
    \label{fig:high-speed}}
\end{figure*}

\subsection{Results and Capabilities}
\noindent \textbf{Fast Motion Recovery:} 
Fig.~\ref{fig:pano-comparison}~(left) shows an example panorama reconstruction in a challenging scenario where the camera moves along an arbitrary trajectory across the full FOV. 
Conventional panorama reconstruction techniques fail, even if there is sufficient light in the scene, because individual frames suffer from extreme motion blur, making it difficult to find reliable feature matches.
By iteratively creating staggered virtual exposures, our method can resolve motion that would otherwise be entirely contained within a single exposure of a conventional camera image. 
Observe that our approach is capable of recovering a near-perfect motion trajectory, which, as seen in the zoomed-in crops, further enables high-fidelity scene reconstruction.

\smallskip
\noindent \textbf{Low Light Robustness:} Fig.~\ref{fig:pano-comparison}~(center) shows the challenging scenario where the camera pans across a dark scene. Here, the conventional RGB method fails because no matches are found in the extremely noisy RGB frames. The situation gets worse in Fig.~\ref{fig:pano-comparison}~(right) where low light is accompanied by extremely fast camera motion. In this extremely low flux regime, the RGB image is dominated by read noise and causes feature registration to fail. In contrast, our approach produces high-quality reconstructions.  

\smallskip
\noindent \textbf{Globally Consistent Matching:} A key issue when global motion is estimated piece by piece is that of drift: any error in the pairwise registration process accumulates over time. This phenomenon is clearly visible in the RGB panorama in Fig.~\ref{fig:pano-comparison}~(left)---not only does the estimated motion trajectory (red) drift away from the ground truth (black), but the panorama gets stretched as compared to the ground truth panorama outline (black dotted line). This drift gets corrected with the proposed method due to the iterative refinement of both the motion estimate and the resulting reconstruction. Fig.~\ref{fig:mulilvl} demonstrates such iterative refinement using real SPAD captures with our hardware prototype. As we increase the number of iterations, the global shape of the panorama gets rectified. The progressive improvement of individual aggregate frames is shown in Fig.~\ref{fig:teaser}(b). \smallskip


\noindent \textbf{Super-Resolution and Efficient Registration:} Due to dense temporal sampling, and the resulting fine-grained homography estimates, the proposed method enables super-resolution in the reconstructed panoramas. This is achieved by applying a scaling transform to the estimated homographies before the merging step. This scaling transform stretches the grid of pixels into a larger grid, resulting in super-resolution. Further, to save on compute and memory costs, this scaling factor can be gradually introduced across iterations. For example, if the goal is to super-resolve by a scale of $4 \times$, we could scale the estimated warps by a factor of two over two iterations. It is also possible to use scaling factors that are smaller than one in the initial iterations of the pipeline. This can be done to create large-scale panoramas, such as the one in Fig.~\ref{fig:teaser}(c), while maintaining low computational and memory footprints. An experimental result with sub-pixel registration is shown in Fig.~\ref{fig:sr}. \smallskip

\begin{figure}[!t]
    \centering \includegraphics[width=0.9\columnwidth]{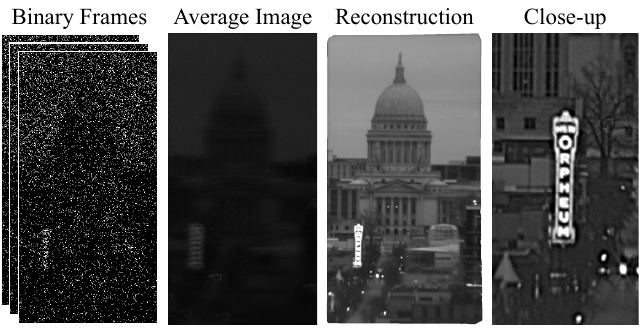}
    \caption{\textbf{High dynamic range image stabilization:} 
    By aligning a large number of extremely dark binary frames, we can stabilize the high-frequency camera shake which causes the average image to be washed out, and faithfully reconstruct this night-time scene, recovering detail in both the dark and bright regions. 
    \label{fig:hdr}}
\end{figure}

\noindent \textbf{High Dynamic Range:} Single photon cameras have recently been demonstrated to have high dynamic range (HDR) capabilities~\cite{ingle_high_2019,ingle_passive_2021,antolovic_dynamic_2018}. By performing high-accuracy homography estimation and registration, the proposed method is able to merge a large number of binary measurements from a given scene point, thus achieving HDR. Fig.~\ref{fig:hdr} shows a real-world example of HDR on a sequence of binary frames captured at night. \smallskip

\noindent \textbf{Extension to High-Speed Cameras:} The stratified re-sampling approach can be extended to other high-speed imaging modalities that allow fast sampling.
The only assumption is that individual frames contain minimal motion blur and can be combined in such a way that boosts SNR and allows for feature matching. 
We demonstrate this using a commercially available high-speed camera (Photron Infinicam) in Fig.~\ref{fig:high-speed}(a). This camera captures $\sim 1000$~fps at its full resolution of $1246 \times 1024$, with higher frame rates available for lower resolutions. All frames are compressed on the camera and access to raw frames is not possible.
If the scene is too dark all useful information will be corrupted by compression artifacts, further impeding the creation of virtual exposures as compression and frame aggregation do not commute.
This phenomenon can be seen in Fig.~\ref{fig:high-speed}(b), it occurs when using the same optical setup as the one used with our SPAD prototype ($75$mm focal length, $f/5.6$), even with a much longer exposure time (which corresponds to $500$fps). 
To get sufficient signal to overcome these limitations we increased the aperture to allow the camera to capture $4\times$ more light.
Fig.~\ref{fig:high-speed}(c) shows a sample frame captured at $500$fps with this new setup.
Despite the slower motion ($\sim 2 \times$ slower than seen in Fig.~\ref{fig:teaser}(c)) and the much larger resolution of the Infinicam, scene details such as text appear blurred.
These frames can be assembled into a larger panorama using our algorithm (Fig.~\ref{fig:high-speed}(c)), albeit, with some residual blur.

Although high-speed sampling enables the creation of virtual exposures, the reconstruction quality deteriorates in challenging conditions due to both the high read noise and the relatively lower sampling rate of high-speed cameras, thus suggesting that the proposed techniques are ideally suited for single-photon imaging. 

\begin{figure}[!t]
    \centering \includegraphics[width=0.9\columnwidth]{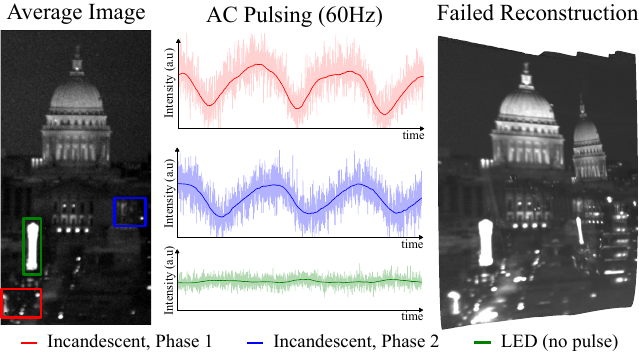}
    \caption{\textbf{Artificial flicker, a failure case:} The flickering of street lights due to the power grid's alternating current can be observed when using high frame rates. This flickering violates the brightness constancy assumption used by most registration algorithms and thus leads to a reconstruction failure.
    \vspace{-0.1in}
    \label{fig:acgrid-failure}}
\end{figure}

\section{Limitations and Future Outlook}\label{sec:discussion}

\begin{figure}[!t]
    \centering \includegraphics[width=0.95\columnwidth]{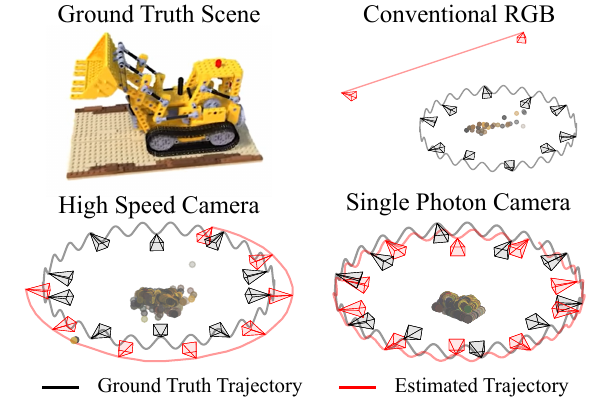}
    \caption{\textbf{High-speed 3D pose estimation using COLMAP:} Naively averaging adjacent binary frames from a single photon camera enables better 3D reconstruction and pose recovery than what is possible with a high-speed or conventional camera. A promising future research direction is to apply the stratified re-sampling ideas put forward in this paper to pose estimation. 
    \label{fig:pose}}
    \vspace{-0.1in}
\end{figure}


\noindent\textbf{Brightness Constancy Failure}: As seen in Fig.~\ref{fig:acgrid-failure}, our hardware prototype is fast enough to detect and even measure the flickering of artificial lighting due to the electric grid. While this can be used to categorize light sources and measure the grid's load~\cite{sheinin2017computational}, it can also cause the underlying registration algorithm to fail as the brightness constancy assumption is violated. This may be mitigated using spatially varying virtual exposures (longer exposures in regions with non-constant brightness). \smallskip

\noindent\textbf{Beyond the Planar Scene Assumption}: 
The applications shown in this paper use a homography-based motion model which characterizes global camera motion under the planar scene assumption. An important next step is to extend these ideas to 3D scenes with global motion, such as 6-DoF pose estimation. Fig.~\ref{fig:pose} shows that the high temporal sampling provided by using single photon cameras improves COLMAP's~\cite{schonberger_structure--motion_2016} pose estimation and sparse reconstruction when simply averaging neighboring frames. How can this initial pose estimate be used to refine our reconstruction? We discuss two possible 3D-consistent aggregation methods below which are promising future research directions. 




\smallskip
\noindent\textbf{Implicit 3D Representations:}
One way to perform the stratified 3D aggregation of binary frames needed to converge to a high-quality pose estimate would be to adapt the recent work done on implicit representations~\cite{mildenhall_nerf_2020,muller_instant_2022,tancik_nerfstudio_2023} to work with binary images.
However numerous challenges remain such as i) how to adapt the rendering model to non-differentiable image data, ii) how to train and update this representation in an online manner, and iii) how the stochastic nature of binary frames will affect  the creation and refinement of a globally consistent 3D representation.    

\smallskip
\noindent\textbf{Dense Motion Models:}
A more general motion model, such as optical flow, could be applied pixel-wise to allow for robust, 3D-consistent binary frame aggregation without resorting to a three-dimensional representation. A recent work~\cite{ma_quanta_2020} has applied optical flow to binary frames with the goal of reconstructing high-quality images as opposed to recovering 3D structure and motion. This method estimates per-frame optical flow in a single shot, meaning that applying our stratified algorithm in this scenario could lead to improved reconstruction and fine-grain pose estimates.

\clearpage
{\small
    \bibliographystyle{ieee_fullname}
    \bibliography{references}
}

\clearpage
\onecolumn
\renewcommand{\figurename}{Supplementary Figure}
\renewcommand{\figurename}{Suppl. Fig.}
 \renewcommand{\thesection}{S.\arabic{section}}
\renewcommand{\theequation}{S\arabic{equation}}
\setcounter{figure}{0}
\setcounter{section}{0}
\setcounter{equation}{0}
\setcounter{page}{1}

\begin{center}
\huge Supplementary Document for\\
\huge ``Panoramas from Photons'' \\[0.5cm]
\large Sacha Jungerman, Atul Ingle, Mohit Gupta\\
{\tt\small sjungerman@wisc.edu, ingle2@pdx.edu, mohitg@cs.wisc.edu}\\[0.5cm]
(ICCV 2023)
\end{center}
\vspace{1em}

\section{Comparison with One-Shot Motion Compensation Methods}
\begin{figure*}[!h]
    \centering \includegraphics[width=1.0\textwidth]{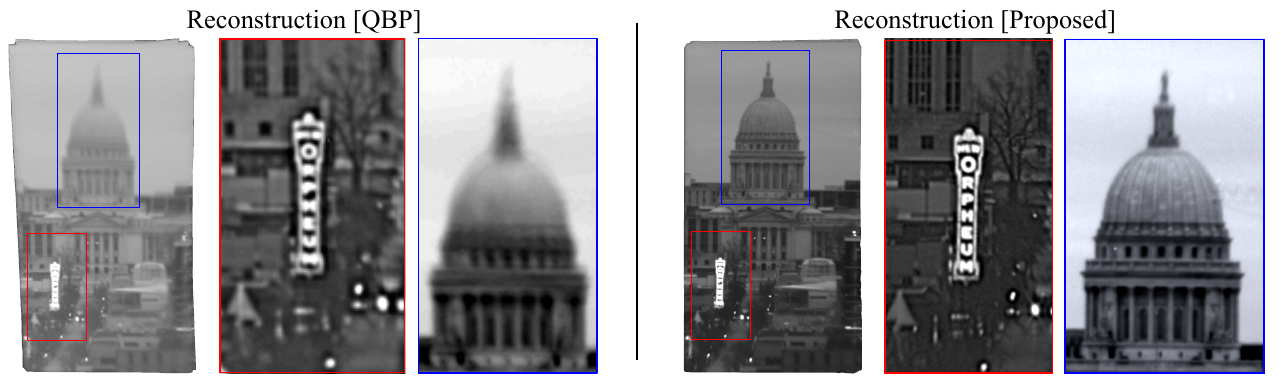}
    \caption{\textbf{Comparison with QBP~\cite{ma_quanta_2020}:} The raw photon data was first processed using QBP and then assembled into the reconstruction shown here using traditional stitching methods. QBP cannot fully compensate for the high-frequency camera vibrations causing the reconstruction to be blurry.     
    \label{supfig:qbp-comparison}}
\end{figure*}



Recent work on motion compensation for single-photon camera data~\cite{ma_quanta_2020,seets_motion_2020,iwabuchi_image_2021} focuses on techniques to aggregate binary frames, in a motion-aware manner, by only making a single pass over the set of binary frames. The stratified method proposed in this paper allows motion to be estimated and refined over multiple iterations, which makes direct comparison with existing techniques difficult. That said, the proposed iterative approach could be used with more general motion models, and therefore, is complementary to existing techniques [24]. An important next step is to apply our method synergistically with these prior works.

Quanta burst photography (QBP)~\cite{ma_quanta_2020} is a recently proposed algorithm that uses a more general optical flow-based motion model that locally warps and registers groups of binary frames.
A comparison with our method is shown in Suppl.~Fig.~\ref{supfig:qbp-comparison}. 
The QBP image is generated by first creating motion-compensated frames from groups of $m=1000$ binary frames. 
These frames are then assembled into the final reconstruction by using a traditional homography estimation technique~\cite{brown_automatic_2007}.
This implementation makes a single pass over the binary frames.
In contrast, our method assumes a planar motion model and uses two iterations to re-estimate the homography warps used for aligning and merging the raw binary frames.
This provides sharper scene details such as the text and smaller features near the top of the dome.\footnote{The QBP image quality is further degraded by quantization artifacts because it operates in the $8$-bit quantized sRGB space whereas our method uses linearized intensity images.}
Both QBP and our method are agnostic to the sequence length and only sensitive to the group size. In fact, many results in the QBP paper [24] use sequences of $10$k frames or less. To ensure fairness of comparison in Suppl.~Fig.~\ref{supfig:qbp-comparison}, we use the \emph{same group size ($m=1000$) and the same total number of frames for QBP and our approach}. Also note that while QBP performs robust merging of binary frames (based on Wiener filtering), our approach simply sums binary frames once warped as we assume the estimated warps are sufficiently accurate. Incorporating this robust merging technique could further reconstruction quality. 


\newpage
\section{Edge Effects and Pre-warping}
\begin{figure*}[!h]
    \centering \includegraphics[width=1.0\textwidth]{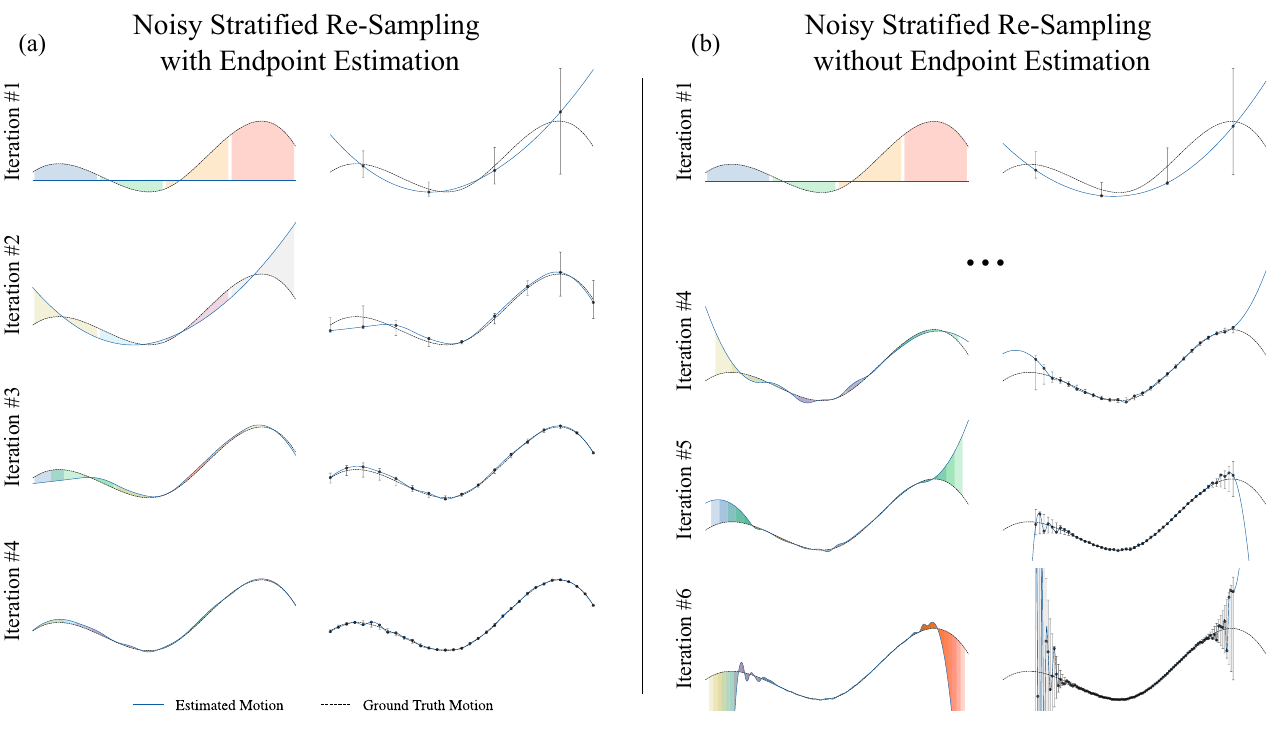}
    \caption{\textbf{Edge Effects in Stratified Temporal Re-Sampling:} (a) The motion estimate provided by the underlying registration algorithm might be noisy (for simplicity, this noise was omitted in Fig.~\ref{fig:iterated-estimation}). Despite this noise, the motion trajectory rapidly converges to the ground truth motion. (b) Over the course of a few iterations, localization errors can accumulate at the ends of the trajectory if the endpoints of the trajectory are not estimated. Thankfully this phenomenon can be mitigated easily by either estimating boundary frames or simply stopping the iterative process before it occurs, as most sequences will converge to a satisfactory motion estimate in two or three iterations. 
    \label{supfig:edge-effects}}
\end{figure*}

When an off-the-shelf homography estimation algorithm is used over a virtual exposure of duration $\tau$, with respect to which time instant within the exposure duration will the estimated localization be? This is not an issue for small camera movements, but with fast motion, this ambiguity has a compounding effect. A sensible assumption would be to presume that the base model estimates the average location over an exposure time, or perhaps, the location at the center of the exposure. This observation is critical for two reasons: i) if we have already compensated for some motion when creating the aggregate frame, our new estimate will be relative to it, and ii) it allows us to localize with respect to any time instant during the exposure by warping the photon data, before aggregation, such that the time instant of interest is warped by the identity warp instead of the current motion estimate.   

Without accounting for the former, any motion estimate would rapidly drift away. In practice, it is beneficial to compensate for this relative offset before aggregation and localization as it helps constrain the size of the aggregate frames and can lead to better matches. The latter enables the localization of off-center time slices of a virtual exposure, enabling precise localization at the boundaries of a captured sequence, which, as seen in Supp. Fig.~\ref{supfig:edge-effects}, is necessary for proper convergence of the motion estimate, and finer motion estimation.

\end{document}